\pdfoutput=1

\documentclass[11pt]{article}

\usepackage[]{acl}

\usepackage{times}
\usepackage{latexsym}

\usepackage[T1]{fontenc}

\usepackage[utf8]{inputenc}

\usepackage{microtype}

\usepackage{algorithm}
\usepackage{microtype}
\usepackage{algcompatible,amsmath}
\usepackage{graphicx}
\usepackage{url}
\usepackage{multirow}
\usepackage{comment}
\usepackage{amssymb}
\usepackage{pifont}
\usepackage{balance}
\usepackage{mathtools}
\usepackage{xcolor}
\usepackage{colortbl}
\usepackage{bbm}
\usepackage{enumitem}
\usepackage{textcomp}
\usepackage{array}
\usepackage{booktabs}
\usepackage{tabularx}
\usepackage{CJKutf8}
\usepackage{newfloat}
\usepackage{listings}
\usepackage{framed}
\usepackage{xcolor}
\usepackage{soul}
\usepackage{listings}
\usepackage{bbding}
\usepackage{makecell}

\usepackage{newfloat}
\usepackage{listings}
\floatstyle{ruled}
\newfloat{listing}{tb}{lst}{}
\floatname{listing}{Listing}

\newcommand{\squishlist}{
\begin{list}{$\bullet$}
{ \usecounter{Lcount}
\setlength{\itemsep}{0pt}
\setlength{\parsep}{0pt}
\setlength{\topsep}{0pt}
\setlength{\partopsep}{0pt}
\setlength{\leftmargin}{2em}
\setlength{\labelwidth}{1.5em}
\setlength{\labelsep}{0.5em} } }

\makeatletter
\newcommand{\thickhline}{%
    \noalign {\ifnum 0=`}\fi \hrule height 2pt
    \futurelet \reserved@a \@xhline
}

\newcommand{\squishend}{
\end{list} }
\newcommand*\circled[1]{\kern-2.5em%
  \put(0,4){\color{white}\circle*{18}}\put(0,4){\circle{10}}%
  \put(-3,0){\color{black}\bfseries#1}~~}
\usepackage{tikz} 
\newcommand*\circledd[1]{\tikz[baseline=(char.base)]{
            \node[shape=circle,draw,inner sep=.8pt] (char) {\bfseries#1};}}
\usepackage{enumitem}
\newcommand{\modelname}{\textsc{\textsf{EnTDA}}}
\newcommand{\xmark}{\ding{55}}%

\title{Entity-to-Text based Data Augmentation for various \\Named Entity Recognition Tasks}

\author{Xuming Hu$^{1}$\thanks{*Work done during an internship at Alibaba DAMO Academy.} , Yong Jiang$^{2}$, Aiwei Liu$^1$, Zhongqiang Huang$^{2}$, Pengjun Xie$^{2}$,\\ \textbf{Fei Huang}$^{2}$, \textbf{Lijie Wen}$^{1}$, \textbf{Philip S. Yu}$^{1,3}$\\
  $^1$Tsinghua University, $^2$Alibaba DAMO Academy, $^3$University of Illinois at Chicago\\
  \texttt{\{hxm19,liuaw20\}@mails.tsinghua.edu.cn}\\
  \texttt{\{yongjiang.jy,chengchen.xpj\}@alibaba-inc.com},
  \\  \texttt{wenlj@tsinghua.edu.cn,psyu@uic.edu}
  }

\begin{document}
\maketitle
\begin{abstract}

Data augmentation techniques have been used to alleviate the problem of scarce labeled data in various NER tasks (flat, nested, and discontinuous NER tasks). Existing augmentation techniques either manipulate the words in the original text that break the semantic coherence of the text, or exploit generative models that ignore preserving entities in the original text, which impedes the use of augmentation techniques on nested and discontinuous NER tasks. In this work, we propose a novel \textit{Entity-to-Text} based data augmentation technique named {\modelname} to add, delete, replace or swap entities in the entity list of the original texts, and adopt these augmented entity lists to generate semantically coherent and entity preserving texts for various NER tasks. Furthermore, we introduce a diversity beam search to increase the diversity during the text generation process. Experiments on thirteen NER datasets across three tasks (flat, nested, and discontinuous NER tasks) and two settings (full data and low resource settings) show that {\modelname} could bring more performance improvements compared to the baseline augmentation techniques.


\end{abstract}
\section{Introduction}
\label{sec:intro}
\begin{figure}[t!]
    \centering
    \includegraphics[width=0.99\linewidth]{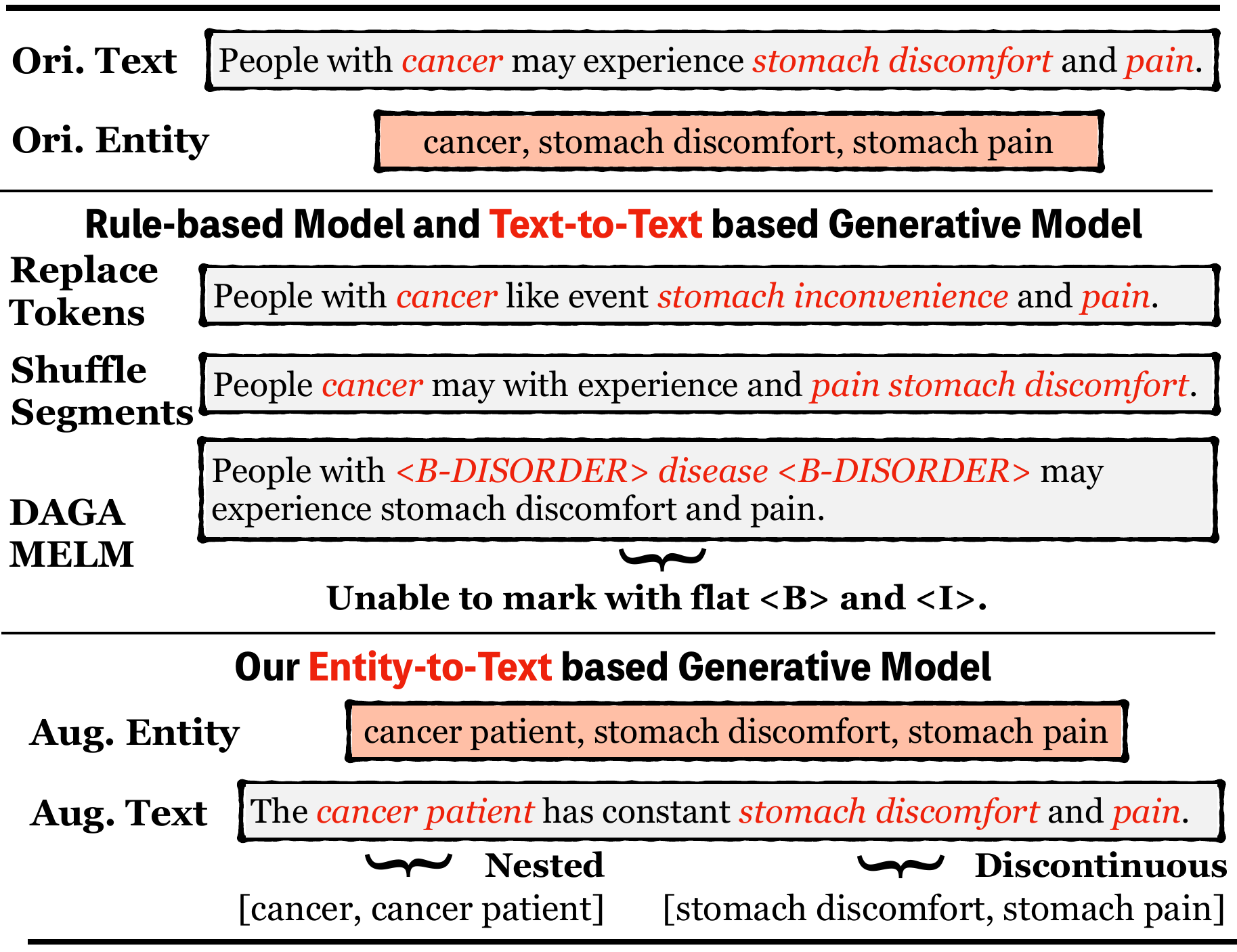}
    \caption{Comparison of augmented cases generated by Rule-based model and \textit{Text-to-Text} based generative model vs. Our \textit{Entity-to-Text} based generative model.}
    \label{fig:example}
\vspace{-4mm}
\end{figure}
Recent neural networks show decent performance when a large amount of training data is available.
However, these manually labeled data are labor-intensive to obtain. Data augmentation techniques \cite{shorten2019survey} expand the training set by generating synthetic data to improve the generalization and scalability of deep neural networks, and are widely used in NLP \cite{feng2021survey,li2022data}.
One successful attempt for data augmentation in NLP is manipulating a few words in the original text, such as word swapping \cite{csahin2018data,min2020syntactic} and random deletion \cite{kobayashi2018contextual,wei2019eda}. These methods generate synthetic texts effortlessly without considering the semantic coherence of sentences.
More importantly, these augmentation approaches work on sentence-level tasks like classification but cannot be easily applied to fine-grained and fragile token-level tasks like Named Entity Recognition (NER).

Named Entity Recognition aims at inferring a label for each token to indicate whether it belongs to an entity and classifies entities into predefined types. Due to transformations of tokens that may change their labels, \citet{dai-adel-2020-analysis} augment the token-level text by randomly replacing a token with another token of the same type. However, it still inevitably introduces incoherent replacement and results in syntax-incorrect texts. DAGA \cite{ding-etal-2020-daga} and MELM \cite{zhou2022melm} investigate the Text-to-Text data augmentation technique using generative methods that preserve semantic coherence and recognize entities through entity tagging during text generation.
However, since it is difficult to use flat $\langle B-Type\rangle$ and $\langle I-Type\rangle$ labels to mark nested and discontinuous entities during text generation, these methods can only be used for flat NER tasks. In addition, only the entities are masked during the generation process, so that the diversity of generated texts is also limited. For example, as shown in Figure \ref{fig:example}, rule-based models replace tokens or shuffle segments, such as ``with'' and ``cancer may'' are shuffled, which makes the augmented text no longer semantically coherent, and even modifies the semantic consistency of the text to affect the prediction of entity labels. The Text-to-Text based generative models cannot leverage flat $\langle B-Type\rangle$ and $\langle I-Type\rangle$ labels to mark the ``stomach'' token in the discontinuous entities: ``stomach discomfort'' and ``stomach pain'', thus limiting the application of this method to nested and discontinuous NER tasks.

\begin{figure*}[t!]
    \centering
    \includegraphics[width=0.94\linewidth]{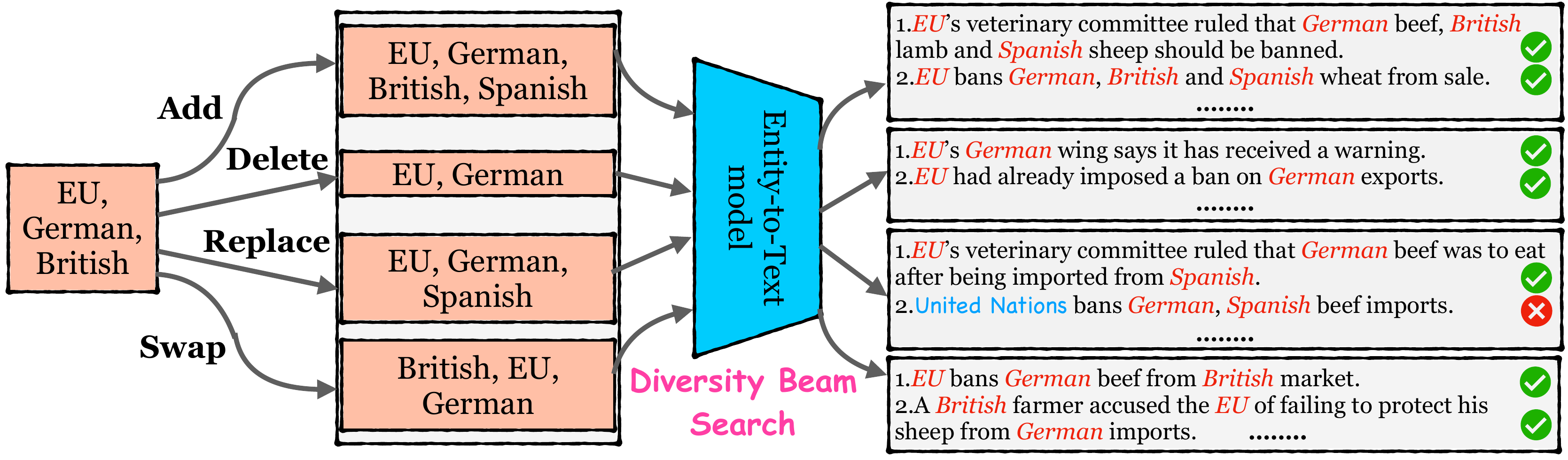}
    \caption{Overview of the proposed Entity-to-Text based data augmentation approach {\modelname}. We first augment entity list via adding, deleting, replacing and swapping entities. Then the augmented entities will generate texts by adopting pretrained language model with diversity beam search. We finally mark the preserved entities in the augmented texts. Note that the texts that do not match the preserved entity list will be discarded.} 
    \label{fig:overview}
\vspace{-0.1in}
\end{figure*}

To maintain text semantic coherence during augmentation and preserve entities for various NER tasks, in this work, we propose a novel \textbf{Entity-to-Text} instead of \textbf{Text-to-Text} based data augmentation approach named {\modelname}. As illustrated in Figure \ref{fig:overview}, we first obtain the entity list [EU, German, British] in the original text, and then add, delete, swap, and replace the entity in the entity list to obtain the augmented entity list, e.g. [EU, German, British, Spanish]. We investigate that leveraging the rule-based methods to modify the entities in the entity list could generate more combinatorial entity lists without introducing grammatical errors. Then we adopt a conditional language model to generate the semantically coherent augmented text based on the augmented entity list. Thanks to the augmented entity list (including flat, nested, and discontinuous entities) we have already obtained, we can mark these preserved entities in the augmented text as shown in Figure \ref{fig:illustration}. Since the augmented entity list provide the 
similar entity information in the text augmented by the language model, which may leads to insufficient diversity of text generation. Therefore, we propose a diversity beam search method for generative models to enhance text diversity.
Overall, the main contributions of this work are as follows:
\squishlist
\item To the best of our knowledge, we propose the first Entity-to-Text based data augmentation technique {\modelname}. {\modelname} leverages the pretrained large language model with semantic coherence and entity preserving to generate the augmented text, which could be used to benefit for all NER tasks (flat, nested, and discontinuous NER tasks).
\item We propose the diversity beam search strategy for {\modelname} to increase the diversity of the augmented text during generation process. 
\item We show that {\modelname} outperforms strong data augmentation baselines across three NER tasks and two settings (full data and low resource settings).
\squishend

\begin{table}
\centering
\scalebox{0.56}{
\begin{tabular}{clccccc}
\thickhline
\multicolumn{2}{c}{\multirow{2}{*}{Techniques}} & \multicolumn{1}{c}{\multirow{2}{*}{Coher.}}& 
\multicolumn{1}{c}{\multirow{2}{*}{Diver.}}& \multicolumn{3}{c}{NER Tasks} 
\\ \cmidrule(lr){5-7} 
& & & & Flat & Nested & Discon.  \\
\midrule

\multicolumn{1}{c}{\multirow{4}{*}{\makecell[c]{Rule \\ Based\\ Techniques}}} &Label-wise token rep. & -- & -- & \Checkmark & \Checkmark& \Checkmark \\ 
&Synonym replacement & -- & -- & \Checkmark & \Checkmark& \Checkmark \\ 
&Mention replacement & -- & -- & \Checkmark & \Checkmark& \xmark\\ 
&Shuffle within segments & -- & -- & \Checkmark & \Checkmark& \Checkmark \\ 

\midrule
\multicolumn{1}{c}{\multirow{3}{*}{\makecell[c]{Generative \\ Techniques}}} &DAGA \cite{ding-etal-2020-daga} & \Checkmark & -- & \Checkmark & \xmark& \xmark \\ 
&MELM \cite{zhou2022melm} & \Checkmark & -- & \Checkmark &\xmark & \xmark \\ 
\cmidrule(lr){2-7} 
&\textbf{\modelname} & \Checkmark & \Checkmark & \Checkmark & \Checkmark& \Checkmark \\ 

\thickhline
\end{tabular}}
\caption{Comparison of different categories of techniques. ``Coher.'' means ``Semantic Coherence'' and ``Diver.'' means ``Diveristy''.}
\label{tab:comparsion}
\vspace{-4mm}
\end{table}

\section{Related Work}
\label{sec:related}

\subsection{Various NER Tasks}
Named Entity Recognition (NER) is a pivotal task in IE which aims at locating and classifying named entities from texts into the predefined types such as \texttt{PERSON}, \texttt{LOCATION}, etc. \cite{chiu2016named,xu2017local,yu2020named}. In addition to flat NER task \cite{sang2003introduction}, \citet{kim2003genia} proposed nested NER task in the molecular biology domain. For example, in the text: \textit{Alpha B2 proteins bound the PEBP2 site}, the entity \textit{PEBP2} belongs to the type \texttt{PROTEIN} and \textit{PEBP2 site} belongs to \texttt{DNA}. 

Furthermore, some entities recognized in the text could be discontinuous \cite{Mowery2013Task1S, Mowery2014Task2S, karimi2015cadec}. For example, in the text: \textit{I experienced severe pain in my left shoulder and neck}, the entities \textit{pain in shoulder} and \textit{pain in neck} contain non-adjacent mentions. Some previous works proposed the unified frameworks which are capable of handling both three NER tasks \cite{li-etal-2020-unified,yan-etal-2021-unified-generative,li-etal-2021-span}. However, there is no unified data augmentation method designed for all three NER tasks due to the complexity of entity overlap. In this work, we try to bridge this gap and propose the first generative augmentation approach {\modelname} that can be used to generate augmented data for all NER tasks (flat, nested, and discontinuous NER tasks).

\subsection{Data Augmentation for NLP and NER}
As shown in Table \ref{tab:comparsion}, we compare {\modelname} with rule-based and traditional generative techniques, and present the comparison results below.
\paragraph{Rule-based Augmentation}
Various rule-based augmentations for NLP tasks such as word replacement \cite{zhang2015character,cai2020data}, random deletion \cite{kobayashi2018contextual,wei2019eda}, and word swapping \cite{csahin2018data,min2020syntactic} manipulate the words in the original texts to generate synthetic texts.
However, these manipulated tokens could not maintain the original labels since the change of syntax and semantics.

\citet{dai-adel-2020-analysis} proposes a replacement augmentation method to decide whether the selected token should be replaced by a binomial distribution, and if so, then the token will be replaced by another token with the same label. Furthermore, the similar approaches could be extended from token-level to mention-level.
However, these methods still inevitably introduce incoherent replacement. In this work, we try to introduce the Entity-to-Text based augmentation approach to improve the coherence of the augmented texts.

\paragraph{Generative Augmentation}
Classic generative augmentations for NLP tasks such as back translation, which could be used to train a question answering model \cite{yu2018qanet} or transfer texts from a high-resource language to a low-resource language \cite{hou2018sequence,xia2019generalized}. 
\citet{anaby2020not,kumar2020data} adopt language model which is conditioned on sentence-level tags to modify original data for classification tasks exclusively. 
To utilize generative augmentation on more fine-grained and fragile token-level NER tasks, \citet{ding-etal-2020-daga} treats the NER labeling task as a text tagging task and requires generative models to annotate entities during generation. \citet{zhou2022melm} builds the pre-trained masked language models on corrupted training sentences and focuses on entity replacement. However, these methods rely on the Text-to-Text based generative models which cannot tag a token with nested labels during generation. In this work, we adopt the Entity-to-Text based generative model to tackle all NER tasks and bootstrap the diversity of the model with diversity beam search.

\section{General NER Task Formulation}\label{NER Tasks}
Considering that {\modelname} has sufficient augmentation ability on flat, nested and discontinuous NER, we first formulate the general NER task framework as follows. Given an input text ${X = [x_{1}, x_{2}, ..., x_{n}]}$ of length ${n}$ and the entity type set ${T}$, the output is an entity list ${E = [{e}_{1}, {e}_{2}, ..., {e}_{m}, ..., {e}_{l}]}$ of ${l}$ entities, where ${{e}_{m}=\left[s_{m1}, d_{m1}, ..., s_{mj}, d_{mj}, t_{m}\right]}$. The ${s,d}$ are the start and end indexes of a space in the text ${X}$. The ${j}$ indicates that the entity consists of ${j}$ spans. The $t_{m}$ is an entity type in the entity type set ${T}$. 
For example, the discontinuous entity \texttt{stomach pain} in the text: ``\textit{The cancer patient has constant \underline{stomach} discomfort and \underline{pain}}'' will be represented as ${{e}_{m}=\left[5,5,8,8,DISORDER\right]}$.

\section{Proposed Method}
The proposed Entity-to-Text based data augmentation approach {\modelname} consists of three modules: Entity List Augmentation, Entity-to-Text Generation, and Augmented Text Exploitation. Now we give the details of the three modules.
\label{sec:baseline}

\subsection{Entity List Augmentation}
Entity List Augmentation aims to adopt four rule-based methods: Add, Delete, Replace, and Swap to modify the entities in the entity list obtained from the original sentences. Now, we give the details of four operations on the original entity list ${E = [{e}_{1}, {e}_{2}, ..., {e}_{m}, ..., {e}_{l}]}$ as follows:
\begin{enumerate}[label=\protect\circled{\arabic*}]
    \item \textbf{Add}. We first randomly select an entity ${{e}_{m}}$ from the entity list ${E}$. Then we search for other entities in the training set and add ${{e}_{m}^{'}}$ with the same entity type as ${{e}_{m}}$ to the original entity list: ${E = [{e}_{1}, {e}_{2}, ..., {e}_{m}, {e}_{m}^{'}, ..., {e}_{l}]}$.
    \item \textbf{Delete}. We randomly select an entity ${{e}_{m}}$ from the original entity list ${E}$ and delete it as ${E = [{e}_{1}, {e}_{2}, ..., {e}_{m-1}, {e}_{m+1},..., {e}_{l}]}$.
    \item \textbf{Replace}. We first randomly select an entity ${{e}_{m}}$ from the original entity list ${E}$. Similar to  \circledd{1}, we search ${{e}_{m}^{'}}$ with the same entity type to replace ${{e}_{m}}$ as ${E = [{e}_{1}, {e}_{2}, ..., {e}_{m}^{'}, ..., {e}_{l}]}$.
    \item \textbf{Swap}. We randomly select two entities ${{e}_{m}}$, ${{e}_{m}^{'}}$ in the original entity list ${E}$ and swap their positions as ${E = [{e}_{1}, {e}_{2}, ..., {e}_{m}^{'},}$
    ${ ...,{e}_{m},..., {e}_{l}]}$.
\end{enumerate}

\subsection{Entity-to-Text Generation}
After we obtain the augmented entity lists, the Entity-to-Text Generation module aims to generate the text for each entity list. Since the augmented entity list provide the similar entity information for augmented text, so we propose a diversity beam search method to increase text diversity.

\begin{figure}
    \centering
    \includegraphics[width=0.98\linewidth]{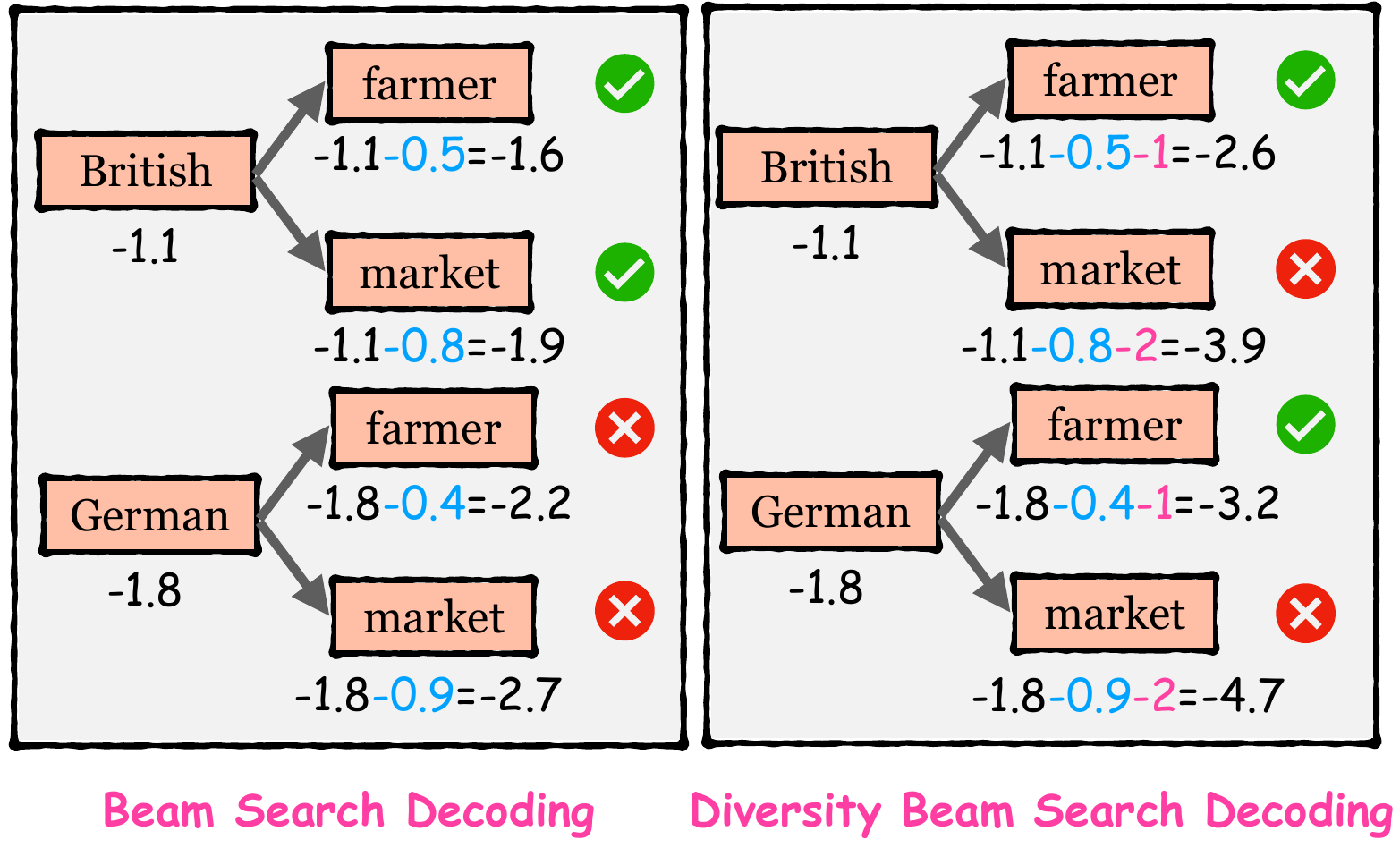}
    \caption{The example about the generated text with beam search decoding (left) and diversity beam search decoding (right). The hyperparameter $\gamma$ is set to 1 and beam width $B$ is set to 2 here.}
    \label{fig:DBS}
\vspace{-4mm}
\end{figure}

Compared to traditional generation models that rely on greedy decoding \cite{chickering2002optimal} and choosing the highest-probability logit at every generation step, we adopt a diversity beam search decoding strategy. More specifically, we first inject the entity types into the augmented entity list $E = [[{t}_{1}], {e}_{1}, [/{t}_{1}],...,[{t}_{m}],{e}_{m},[/{t}_{m}], ...,[{t}_{l}], {e}_{l},[/{t}_{l}]]$
as the input sequence, which should provide sufficient type guidance for the generation model, then we adopt T5 \cite{raffel2020exploring} as the generation model. We first fine-tune T5 on the original Entity-to-Text data and then adopt T5 ($\theta$) to estimate the conditional probability distribution over all tokens in the dictionary $\mathcal{V}$ at time step $t$ as:
\begin{align}\label{eq:pro}
\theta\left(y_{t}\right)=\log \operatorname{Pr}\left(y_{t} \mid y_{t-1}, \ldots, y_{1}, E\right).
\end{align}
where ${y_{t}}$ is the ${{t}^{th}}$ output token ${y}$ in texts. We simplify the sum of $\log$-probabilities (Eq. \ref{eq:pro}) of all previous tokens decoded $\Theta(\mathbf{y}_{[t]})$ as:
\begin{align}\label{eq:decoding}
\Theta\left(\mathbf{y}_{[t]}\right)=\sum_{\tau \in[t]} \theta\left(y_{\tau}\right),
\end{align}
where $\mathbf{y}_{[t]}$ is the token list consisting of $[y_{1},y_{2},...,y_{t}]$. Therefore, our decoding problem is transformed into the task of finding the text that could maximize $\Theta(\mathbf{y})$. The classical approximate decoding method is the beam search \cite{wiseman2016sequence}, which stores top beam width $B$ candidate tokens at time step $t$. Specifically, beam search selects the $B$ most likely tokens from the set:
\begin{align}
\mathcal{Y}_{t}=Y_{[t-1]} \times \mathcal{V},
\end{align}
where $Y_{[t-1]}=\left\{\mathbf{y}_{1,[t-1]}, \ldots, \mathbf{y}_{B,[t-1]}\right\}$ and $\mathcal{V}$ is the dictionary. However, traditional beam search keeps a small proportion of candidates in the search space and generates the texts with minor perturbations \cite{huang2008forest}, which impedes the diversity of generated texts. Inspired by \citet{vijayakumar2016diverse}, we introduce an objective to increase the dissimilarities between candidate texts and finalize the Eq. \ref{eq:decoding} as diversity beam search decoding:
\begin{align}
\hat{\Theta}\left(\mathbf{y}_{[t]}\right)=\sum_{\tau \in[t]} (\theta\left(y_{\tau}\right) - \gamma k_{\tau}),
\end{align}
where $\gamma$ is a hyperparameter and represents the punishment degree. $k_{\tau}$ denotes the ranking of the current tokens among candidates. In practice, it's a penalty text of beam width: $[1,2,...,B]$ which punishes bottom ranked tokens among candidates and thus generates tokens from diverse previous tokens. For a better understanding, we give an example about the text with beam search decoding and diversity beam search decoding in Figure \ref{fig:DBS}. 

The traditional greedy decoding chooses the highest-probability logit at every generation step and results in \textit{British farmer}. Compared to the diversity beam search decoding method, the beam search decoding method maintains a small proportion of candidates in the search space without the introduction of a penalty text of beam width $[1,2,...,B]$. This additional objective increases the dissimilarities between candidate texts and thus generates tokens from diverse previous tokens. For example, \textit{British farmer} and \textit{German farmer} are generated instead of \textit{British farmer} and \textit{British market}, which brings the diversity token \textit{German}. Likewise, the diversity token \textit{market} will also be considered in the subsequent generation.
Overall, at each time step $t$:
\begin{align}
Y_{[t]}=\underset{\mathbf{y}_{1,[t]}, \ldots, \mathbf{y}_{B,[t]} \in \mathcal{Y}_{t}}{\operatorname{argmax}} \sum_{b \in[B]} \hat{\Theta}\left(\mathbf{y}_{b,[t]}\right).
\end{align}
This process will generate the most likely texts that are selected by ranked the $B$ beams based on the diversity beam search decoding.

\begin{figure}[t!]
    \centering
    \includegraphics[width=0.98\linewidth]{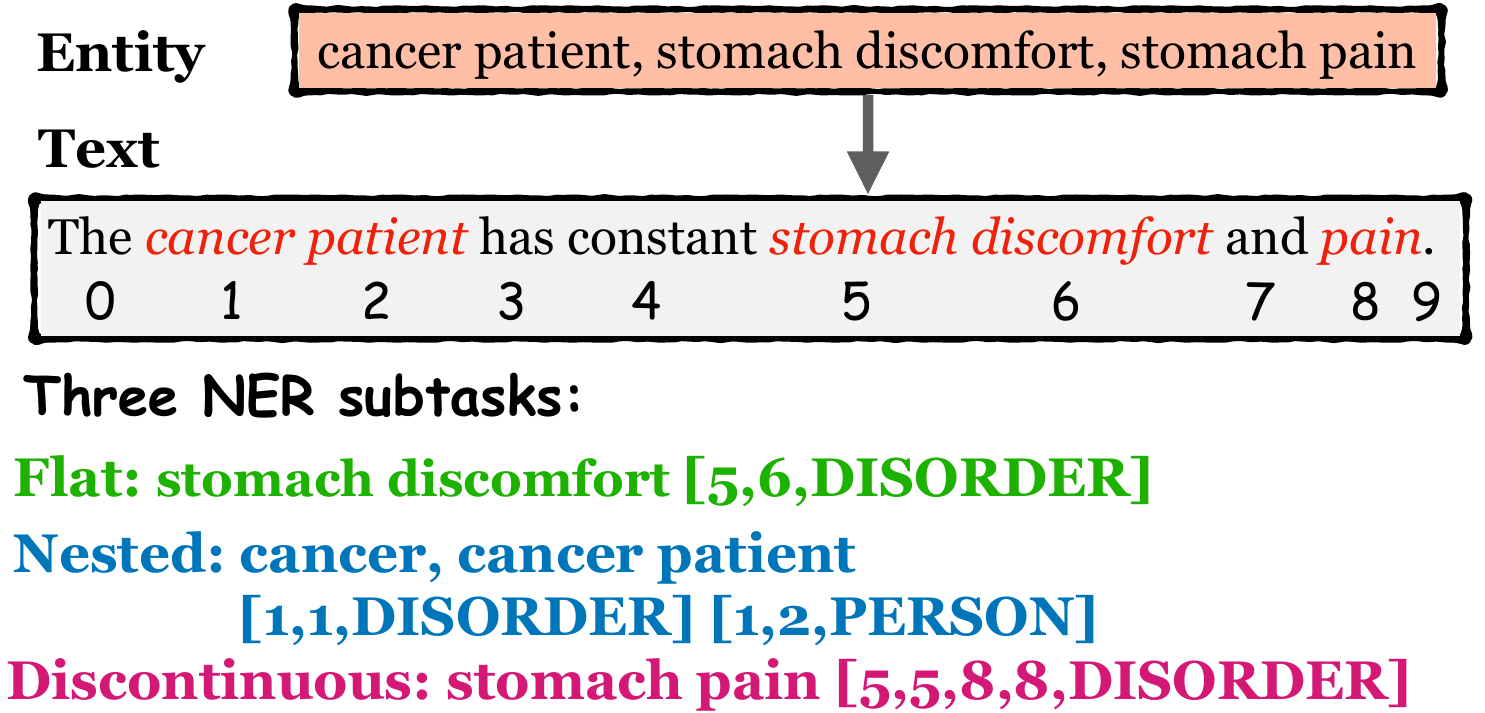}
    \caption{Details of marking the texts with the augmented entity lists for three NER tasks.}
    \label{fig:illustration}
\vspace{-4mm}
\end{figure}

\subsection{Augmented Text Exploitation}\label{exploitation}
To utilize these augmented Entity-to-Text data, we need to mark the texts with the augmented entity lists. 
As illustrated in Figure \ref{fig:overview}, we first automatically judge whether the entities match the tokens in the texts and remove these noisy texts of mismatched entities.
For example, \textit{EU} is generated as \textit{United Nations} and this generated text is automatically deleted.
Then as illustrated in Figure \ref{fig:illustration}, we provide the details of the text marking process:

(1) If the entity is \textbf{flat}, we obtain the start and end position indexes through the exact match between entity and text.

(2) If the entity is \textbf{nested}, we first store all the overlapping entity mentions belonging to the same nested entity and match these mentions with text to obtain start and end position indexes.

(3) If the entity is \textbf{discontinuous}, we match the entity mentions which belong to the same discontinuous entity with text to obtain start and end position indexes. 

Note that the process of text marking is done automatically based on the above three situations. After we obtain these augmented data with marked flat, nested, and discontinuous entities, we naturally formulate the texts as input to NER tasks.

\section{Experiments and Analyses}
\label{sec:experiments}

We conduct extensive experiments on thirteen NER datasets across three tasks (flat, nested, and discontinuous NER) and two settings (full data and low resource NER) to show the effectiveness of {\modelname} on NER, and give a detailed analysis.

\subsection{Backbone Models}
We adopt two SOTA backbone models which could solve all three NER tasks: 



\noindent1) \textbf{The unified Seq2Seq framework} \cite{yan-etal-2021-unified-generative} formulates three NER tasks as an entity span text generation task without the special design of the tagging schema to enumerate spans. 

\noindent2) \textbf{The unified Word-Word framework} \cite{li2022unified} models the neighboring relations between entity words as a 2D grid and then adopts multi-granularity 2D convolutions for refining the grid representations.

These two backbone models are leveraged to solve the general NER tasks illustrated in Section \ref{NER Tasks} and demonstrate the effectiveness of {\modelname}.

\begin{table*}[th!]
\centering
\vspace{-1mm}
\scalebox{0.68}{
\begin{tabular}{lcccccccccc}
\thickhline
\multicolumn{1}{c}{\multirow{2}{*}{Method / Datasets}} & \multicolumn{2}{c}{Flat NER datasets} & \multicolumn{3}{c}{Nested NER datasets} & \multicolumn{3}{c}{Discontinuous NER datasets} & \multicolumn{1}{c}{\multirow{2}{*}{AVG.}}  & \multicolumn{1}{c}{\multirow{2}{*}{$\Delta$}} 
\\ \cmidrule(lr){2-3} \cmidrule(lr){4-6} \cmidrule(lr){7-9} 
&CoNLL2003 & OntoNotes & ACE2004 & ACE2005 & Genia & CADEC & ShARe13 & ShARe14  \\
\midrule 
Unified Word-Word Framework  &93.14 &90.66 &87.54 &86.72 &81.34 &73.22 &82.57 &81.79 &84.62 &-- \\
\multicolumn{1}{l}{$+$Label-wise token rep.} &93.32 &90.78 &87.83 &\underline{86.98} &\underline{81.65} &73.47 &82.84 &82.07 &84.87&\color{red}0.25$\uparrow$ \\
\multicolumn{1}{l}{$+$Synonym replacement} &93.35 &90.75 &87.87 &86.93 &81.63 &\underline{73.50} &\underline{82.87} &\underline{82.10} &\underline{84.88} &\color{red}0.26$\uparrow$ \\
\multicolumn{1}{l}{$+$Mention replacement} &93.29 &90.80 &\underline{87.89} &86.97 &81.64 &-- &-- &-- &-- &--\\
\multicolumn{1}{l}{$+$Shuffle within segments} &93.30 &90.68 &87.68 &86.84 &81.47 &73.36 &82.71 &81.92 &84.75 &\color{red}0.13$\uparrow$ \\
\multicolumn{1}{l}{$+$DAGA } & 93.47 &90.89 &-- &-- &-- &-- &-- &-- &-- &--  \\
\multicolumn{1}{l}{$+$MELM } &\underline{93.60} &\underline{91.06} &-- &-- &-- &-- &-- &-- &-- &--  \\
\rowcolor{gray!10}
\multicolumn{1}{l}{\textbf{$+\modelname$ (Delete)}} &93.82	&91.23&	\textbf{88.29}	&87.54	&82.12&	73.86	&83.31	&82.45  &85.33 &\color{red}0.71$\uparrow$ \\
\rowcolor{gray!10}
\multicolumn{1}{l}{\textbf{$+\modelname$ (Add)}} &\textbf{93.93}	&91.26	&88.27&	\textbf{87.60}	&82.19	&\textbf{73.89}	&83.34	&\textbf{82.55} & \textbf{85.42} &\color{red}\textbf{0.76}$\uparrow$ \\
\rowcolor{gray!10}
\multicolumn{1}{l}{\textbf{$+\modelname$ (Replace)}} &93.87	&91.21	&88.18&	87.46	&\textbf{82.40}	&73.82&	83.19	&82.52  &85.33 &\color{red}0.71$\uparrow$ \\
\rowcolor{gray!10}
\multicolumn{1}{l}{\textbf{$+\modelname$ (Swap)}} &93.91	&91.25&	88.18	&87.54	&82.32	&73.81	&83.30	&82.52 &85.35 &\color{red}0.73$\uparrow$ \\
\rowcolor{gray!10}
\multicolumn{1}{l}{\textbf{$+\modelname$ (All)}} &93.88	&\textbf{91.34}  &	88.21	&87.56	&82.25	&73.86&	\textbf{83.35}&	82.47  &85.37 &\color{red}0.75$\uparrow$ \\
\cmidrule{2-11} 
\multicolumn{1}{l}{\textbf{$+\modelname$ (None)}} &93.44 &90.89 &87.84 &87.01 &81.73 &73.57 &82.90 &82.09  &84.93 &\color{red}0.31$\uparrow$ \\
\multicolumn{1}{l}{\textbf{$+\modelname$ (All) w/o Diver.}} &93.55 &91.01 &87.93 &87.23 &81.91 &73.75 &83.02 &82.20  &85.08 &\color{red}0.46$\uparrow$ \\
\midrule 
\midrule 
Unified Seq2Seq Framework &92.78 &89.51 &86.19 &84.74 &79.10 &70.76 &79.69 &79.40 &82.78 &-- \\
\multicolumn{1}{l}{$+$Label-wise token rep.} &92.91 &89.68 &\underline{86.33} &85.04 &79.41 &\underline{71.22} &\underline{79.93} &\underline{79.64} &83.03 &\color{red}0.25$\uparrow$ \\
\multicolumn{1}{l}{$+$Synonym replacement} &92.85 &89.59 &86.28 &\underline{85.32} &79.36 &71.18 &79.86 &79.55 &83.00 &\color{red}0.22$\uparrow$ \\
\multicolumn{1}{l}{$+$Mention replacement} &92.80 &89.80 &86.14 &85.01 &\underline{79.44} &-- &-- &-- &-- &--  \\
\multicolumn{1}{l}{$+$Shuffle within segments } &92.85 &89.40 &86.22 &84.99 &79.28 &71.13 &79.72 &79.50 &82.89 &\color{red}0.11$\uparrow$ \\
\multicolumn{1}{l}{$+$DAGA } &92.92 &\underline{89.97} &-- &-- &-- &-- &-- &-- &-- &--  \\
\multicolumn{1}{l}{$+$MELM } &\underline{92.95} &89.95 &-- &-- &-- &-- &-- &-- &-- &--  \\
\rowcolor{gray!10}
\multicolumn{1}{l}{\textbf{$+\modelname$ (Delete)}} &93.38 &90.23 &86.51 &86.26 &80.80 &71.51 &80.58 &80.04  &83.67 &\color{red}0.89$\uparrow$ \\
\rowcolor{gray!10}
\multicolumn{1}{l}{\textbf{$+\modelname$ (Add)}} &93.27 &90.27 &86.73 &86.39 &80.88 &71.50 &\textbf{80.92} &80.16  &83.77 &\color{red}0.99$\uparrow$ \\
\rowcolor{gray!10}
\multicolumn{1}{l}{\textbf{$+\modelname$ (Replace)}} &93.32 &90.16 &86.55 &\textbf{86.41} &80.74 &71.64 &80.64 &80.23 &83.71 &\color{red}0.93$\uparrow$ \\
\rowcolor{gray!10}
\multicolumn{1}{l}{\textbf{$+\modelname$ (Swap)}} &93.45 &90.04 &86.40 &86.30 &80.67 &71.37 &80.37 &80.12  &83.59 &\color{red}0.81$\uparrow$ \\
\rowcolor{gray!10}
\multicolumn{1}{l}{\textbf{$+\modelname$ (All)}} &\textbf{93.51} &\textbf{90.31} &\textbf{86.92} &86.39 &\textbf{80.94} &\textbf{71.70} &80.83 &\textbf{80.36}  &\textbf{83.87} &\color{red}\textbf{1.09}$\uparrow$ \\
\cmidrule{2-11} 
\multicolumn{1}{l}{\textbf{$+\modelname$ (None)}} &92.90 &90.02 &86.28 &85.57 &79.66 &71.30 &80.13 &79.71  &83.20 &\color{red}0.42$\uparrow$ \\
\multicolumn{1}{l}{\textbf{$+\modelname$ (All) w/o Diver.}} &93.13 &90.21 &86.47 &85.78 &79.88 &71.54 &80.31 &79.97  &83.41 &\color{red}0.63$\uparrow$ \\

\thickhline

\end{tabular}}
\caption{F1 results of various NER tasks. For all three backbone models and six baseline augmentation approaches, we rerun their open source code and adopt the given parameters.}
\label{tab:verification}

\vspace{-4mm}
\end{table*}
\subsection{Datasets}
To demonstrate that {\modelname} could be used in various NER tasks and backbone models, we follow \citet{yan-etal-2021-unified-generative,li2022unified} and adopt the same datasets (split) as follows:

1) \textbf{Flat NER Datasets}: We adopt the CoNLL-2003 \cite{sang2003introduction} and OntoNotes \cite{pradhan2013towards} datasets. For OntoNotes, we evaluate in the English corpus with the same setting as \citet{yan-etal-2021-unified-generative}.

2) \textbf{Nested NER Datasets}: We adopt the ACE 2004 \cite{doddington2004automatic}, ACE 2005 \cite{Walker2005Ace} and GENIA \cite{kim2003genia} datasets. Following \citet{yan-etal-2021-unified-generative}, we split the ACE 2004/ACE 2005 into train/dev/test sets by 80\%/10\%/10\% and GENIA into 81\%/9\%/10\% respectively.

3) \textbf{Discontinuous NER Datasets}
We adopt the CADEC \cite{karimi2015cadec}, ShARe13 \cite{Mowery2013Task1S} and ShARe14 \cite{Mowery2014Task2S} datasets from biomedical domain. Following \citet{yan-etal-2021-unified-generative}, we split the CADEC into train/dev/test sets by 70\%/15\%/15\% and use 10\% training set as the development set for ShARe13/ShARe14.

We show the detailed statistics and entity types of the datasets in Appendix \ref{Dataset}.

\subsection{Baseline Augmentation Methods}
Unlike sentence-level classification tasks, NER is a fine-grained token-level task, so we adopt six entity-level data augmentation baselines, which are designed for various NER tasks.

The four rule-based baseline augmentation techniques: (1) \textbf{Label-wise token replacement} \cite{dai-adel-2020-analysis} utilizes a binomial distribution to decide whether each token should be replaced, and then replaces the chosen token with another token that has the same entity type. (2) \textbf{Synonym replacement} \cite{dai-adel-2020-analysis} replaces the chosen token with the synonym retrieved from WordNet. (3) \textbf{Mention replacement} \cite{dai-adel-2020-analysis} replaces the chosen entity with another entity, which has the same entity type. (4) \textbf{Shuffle within segments} \cite{dai-adel-2020-analysis} splits the sentences into segments based on whether they come from the same entity type, and uses a binomial distribution to decide whether to shuffle tokens within the same segment. The two generative baseline augmentation techniques are: (5) \textbf{DAGA} \cite{ding-etal-2020-daga} treats the NER labeling task as a text tagging task and annotates entities with generative models during generation. (6) \textbf{MELM} \cite{zhou2022melm} generates augmented data with diverse entities, which is built upon pre-trained masked language models. MELM is further finetuned on corrupted training sentences with only entity tokens being randomly masked to focus on entity replacement.

We present another model: \textbf{$\modelname$  (All)}, which adopts four entity list operations simultaneously to generate augmented texts. Note that we focus on entity-level NER augmentation tasks, so to the best of our knowledge, we have employed all entity-level augmentation techniques.

\subsection{Experiment Settings}
For $\modelname$, we fine-tune the T5-Base \cite{raffel2020exploring} with the initial parameters on the Entity-to-Text data of the training set and utilize the default tokenizer with max-length as 512 to preprocess the data. We use AdamW \cite{loshchilov2018fixing} with $5e{-5}$ learning rate to optimize the cross entropy loss. The batch size is set to 5 and the number of training epoch is set to 3. During diversity beam search decoding, we set $\gamma$ as 10 and beam width ${B}$ as 3, which means that each entity set will generate three texts. 

{\modelname} and all baselines augment the training set by 3x for a fair comparison. For example, the number of texts in the training set is 100, we generate 300 texts and add them to the training set. We replace the language model in MELM \cite{zhou2022melm} with XLM-RoBERTa-large (355M) \cite{conneau2020unsupervised}, and we use T5-Base (220M) with fewer parameters for comparison.

\subsection{Results and Analyses}
Table \ref{tab:verification} shows the average F1 results on three runs. All backbone NER models gain F1 performance improvements from the augmented data when compared with the models that only use original training data, demonstrating the effectiveness of data augmentation approaches in the various NER tasks.
Surprisingly, ${\modelname}$ (None) outperforms the baseline methods by 0.11\% F1 performance among the backbone models, which shows that the generative models using a diversity beam search have sufficient capacity to generate high-quality augmented data. 

More specifically, for flat NER datasets, MELM is considered as the previous SOTA data augmentation approach. The proposed ${\modelname}$ (All) on average achieves 0.23\% higher in F1 among flat NER datasets and two backbone models.
For nested and discontinuous NER datasets, 
the label-wise token replacement method achieves the best performance among baselines. ${\modelname}$ (All) achieve an average 0.78\% F1 boost among nested and discontinuous NER datasets, which demonstrates that leveraging generative model to augment semantically coherent texts is effective.
\begin{table}[bt!]
\centering
\resizebox{0.98\linewidth}{!}{
\begin{tabular}{lccc}
\thickhline
Method / Datasets & CoNLL2003  & ACE2005 & CADEC   \\
\midrule 
Unified Word-Word Framework &86.83  &79.56 &65.03\\
\multicolumn{1}{l}{+Label-wise token rep.} &87.23 &79.97 &\underline{65.50} \\
\multicolumn{1}{l}{+Synonym replacement} &87.16 &80.01 &65.46 \\
\multicolumn{1}{l}{+Mention replacement} &87.30 &\underline{80.10} &--\\
\multicolumn{1}{l}{+Shuffle within segments} &87.04 &79.85 &65.28\\
\multicolumn{1}{l}{+DAGA} &87.82 &-- &-- \\
\multicolumn{1}{l}{+MELM} &\underline{88.24} &-- &-- \\
\rowcolor{gray!10}
\multicolumn{1}{l}{\textbf{+$\modelname$ (Delete)}} & 89.91 &81.94 &69.12 \\
\rowcolor{gray!10}
\multicolumn{1}{l}{\textbf{+$\modelname$ (Add)}} &90.13 &\textbf{82.15} &69.03 \\
\rowcolor{gray!10}
\multicolumn{1}{l}{\textbf{+$\modelname$ (Replace)}} &90.07 &82.01 &69.29\\
\rowcolor{gray!10}
\multicolumn{1}{l}{\textbf{+$\modelname$ (Swap)}} &89.97 &81.98 &69.25 \\
\rowcolor{gray!10}
\multicolumn{1}{l}{\textbf{+$\modelname$ (All)}} &\textbf{90.22} &82.08 &\textbf{69.31} \\

\midrule 
\midrule 
Unified Seq2Seq Framework &85.90  &77.32 &62.24 \\
\multicolumn{1}{l}{+Label-wise token rep.} &86.44 &77.81 &62.56 \\
\multicolumn{1}{l}{+Synonym replacement}&86.73 &77.79 &\underline{62.61} \\
\multicolumn{1}{l}{+Mention replacement} &86.94 &\underline{77.83} &-- \\
\multicolumn{1}{l}{+Shuffle within segments} &86.26 &77.65 &62.49 \\
\multicolumn{1}{l}{+DAGA} &87.05 &-- &-- \\
\multicolumn{1}{l}{+MELM} &\underline{87.43} &-- &-- \\
\rowcolor{gray!10}
\multicolumn{1}{l}{\textbf{+$\modelname$ (Delete)}} & 89.20 &79.10 &66.04 \\
\rowcolor{gray!10}
\multicolumn{1}{l}{\textbf{+$\modelname$ (Add)}} &89.62 &79.23 &\textbf{66.42} \\
\rowcolor{gray!10}
\multicolumn{1}{l}{\textbf{+$\modelname$ (Replace)}} &89.41 &79.02 &66.21 \\
\rowcolor{gray!10}
\multicolumn{1}{l}{\textbf{+$\modelname$ (Swap)}} &88.96 &78.96 &65.93 \\
\rowcolor{gray!10}
\multicolumn{1}{l}{\textbf{+$\modelname$ (All)}} &\textbf{89.82} &\textbf{79.51} &66.40 \\

\thickhline
\end{tabular}}
\caption{F1 results of various NER tasks under low resource scenarios.}
\label{tab:low-resource}
\vspace{-4mm}
\end{table}

Among all NER datasets, ${\modelname}$ is undoubtedly capable of achieving state-of-the-art results (with student's T test $p<0.05$). Except ${\modelname}$ (All), {\modelname} (Add) achieves the largest F1 performance gains of 0.99\% and 0.76\% on the unified Seq2Seq and Word-Word frameworks, respectively. We attribute this delightful improvement of the ``Add'' operation to the additionally introduced knowledge: we add the entity from the training set with the same entity type.

\noindent\textbf{Ablation Study}

In Table \ref{tab:verification}, we remove the entity list augmentation module (\modelname (None)), or change the diversity beam search to the traditional beam search (\modelname (All) w/o Diver.). We can conclude that entity list augmentation and diversity beam search modules bring an average F1 improvement of 0.56\% and 0.38\% on the eight datasets. Using the entity list augmentation module can give a richer entity combination, which brings more improvement. Adopting the diversity beam search brings more diverse texts and gains greater improvements.

\noindent\textbf{Handling Low Resource NER Scenarios}

We further introduce an extreme yet practical scenario: only limited labeled data is available. This low resource NER scenario demonstrates that our {\modelname} approach bootstraps the generalization ability of the NER model and is a quite appealing approach for data-oriented applications in the real-world. In practice, we randomly choose 10\% training data from CoNLL2003/ACE2005/CADEC to represent the three NER tasks. Note that the fine-tuning of T5-large and our four operations on the entity list are also done on 10\% training data.

From Table \ref{tab:low-resource}, compared to training directly on the 10\% training set, leveraging the augmented data achieves the performance improvement in F1. We also observe that {\modelname} approach obtains the most competitive F1 performance improvement when compared with baseline data augmentation approaches. 
More specifically, ${\modelname}$ (All) achieve an average 2.97\% F1 boost among three backbone models, which means {\modelname} obtains more performance gains under the low resource scenario than in the full data scenario. Especially for the most challenging discontinuous dataset CADEC, ${\modelname}$ (All) obtains the largest F1 performance gain of 4.22\%.
Surprisingly, on 10\% CoNLL2003, ${\modelname}$ (All) has only a 2.94\% decrease in F1 performance compared to using the full training data, but ${\modelname}$ (All) saves 10x the annotated data, which shows that adopting ${\modelname}$ is quite appealing for real-world applications.
\begin{table}[bt!]
\centering
\resizebox{0.99\linewidth}{!}{
\begin{tabular}{lccccc}
\thickhline
Method / Datasets & Politics & Natural Science & Music & Literature & AI   \\
\midrule 
Seq2Seq Framework &70.11  &70.72 &72.90 &63.69 &56.77 \\
\multicolumn{1}{l}{+Label-wise token rep.} &70.45  &70.91 &73.48 &63.97 &57.04\\
\multicolumn{1}{l}{+Synonym replacement}&70.43  &71.04 &73.66 &63.92 &57.34\\
\multicolumn{1}{l}{+Mention replacement} &70.47  &71.07 &\underline{73.54} &64.02 &57.42 \\
\multicolumn{1}{l}{+Shuffle within segments} &70.39  &70.94 &73.30 &63.88 &57.26 \\
\multicolumn{1}{l}{+DAGA} &\underline{71.06}  &\underline{71.51} &73.46 &\underline{64.21} &\underline{57.83} \\
\rowcolor{gray!10}
\multicolumn{1}{l}{\textbf{+$\modelname$ (Delete)}} & 72.60 &72.05 &75.87 & 67.18 &61.58 \\
\rowcolor{gray!10}
\multicolumn{1}{l}{\textbf{+$\modelname$ (Add)}} &72.81 &\textbf{72.55} &76.20 &67.82 &61.97 \\
\rowcolor{gray!10}
\multicolumn{1}{l}{\textbf{+$\modelname$ (Replace)}} &72.94 &72.46 &76.12 &67.57 &61.89\\
\rowcolor{gray!10}
\multicolumn{1}{l}{\textbf{+$\modelname$ (Swap)}}& 72.47 &71.89 &75.58 & 67.06 &61.37 \\
\rowcolor{gray!10}
\multicolumn{1}{l}{\textbf{+$\modelname$ (All)}}& \textbf{72.98} &72.47 &\textbf{76.55} & \textbf{68.04} &\textbf{62.31} \\
\thickhline
\end{tabular}}
\vspace{-2mm}
\caption{F1 results of real low resource NER tasks.}
\label{tab:real}
\vspace{-2mm}
\end{table}

\noindent\textbf{Tackling Real Low Resource NER Tasks}

We adopt real low resource NER datasets \cite{liu2021crossner} from Wikipedia which contains politics, natural science, music, literature and artificial intelligence domains with only 100 or 200 labeled texts in the training set. {\modelname} and baseline data augmentation approaches still augment the training set by 3x. 
From Table \ref{tab:real}, we are delighted to observe {\modelname} could quickly learn from the extremely limited Entity-to-Text data and bring 3.45\% F1 performance gains over various domains. Compared with baseline augmentation methods, {\modelname} generates more diverse texts and undoubtedly gains greater advantages.

\begin{figure}[t!]
    \centering
    \includegraphics[width=0.99\linewidth]{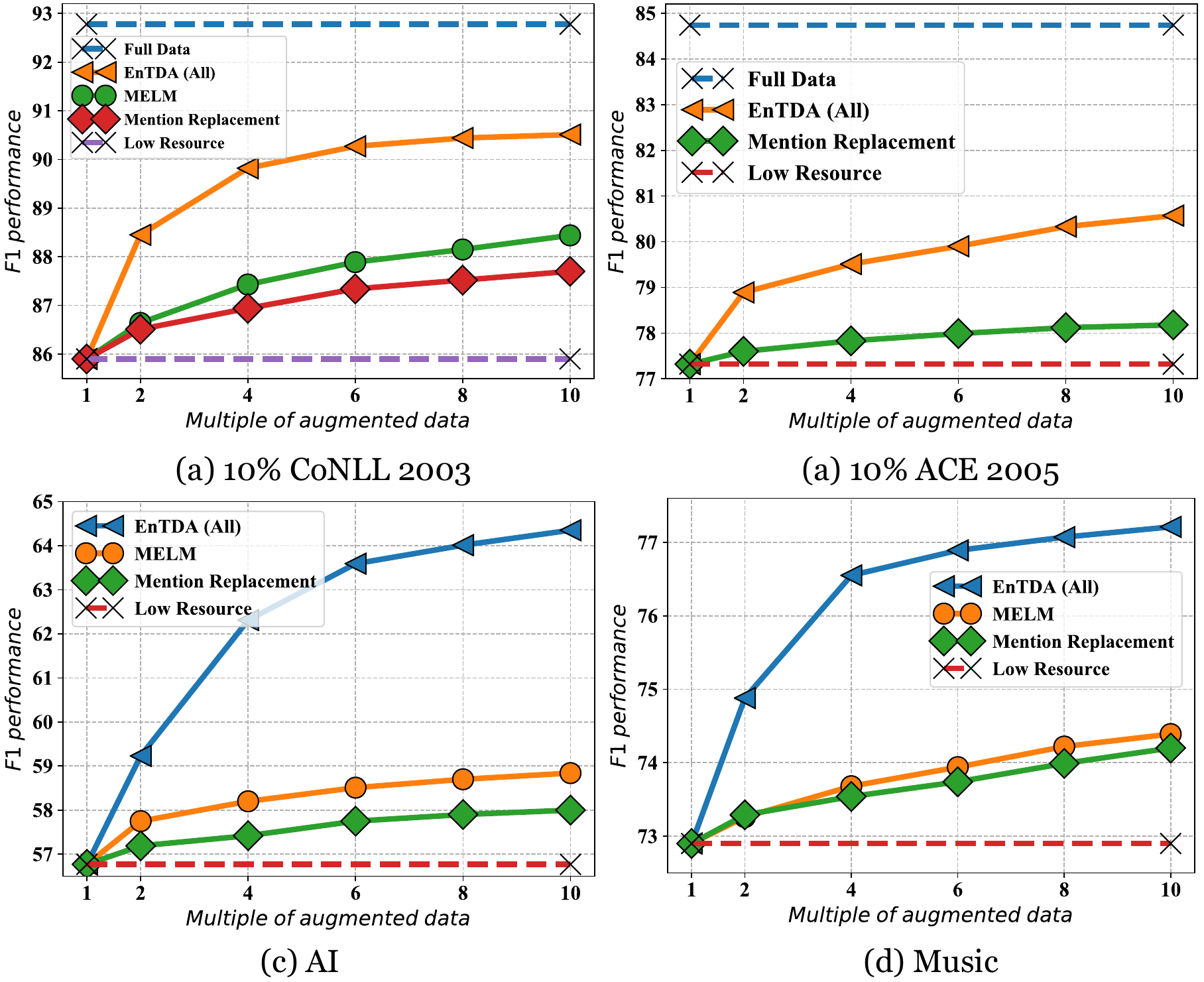}
    \caption{F1 results of the unified Seq2Seq framework with augmented data at various multiples on different low resource datasets.}
    \label{fig:real}
\vspace{-4mm}
\end{figure}


\begin{table}[t!]
\centering
\resizebox{0.9\linewidth}{!}{
\begin{tabular}{lccc}
\thickhline
Method / Datasets & CoNLL2003 & CADEC  & AI   \\
\midrule 
\multicolumn{1}{l}{Label-wise token rep.} &8.12  &8.87 &7.52 \\
\multicolumn{1}{l}{Synonym replacement}& 7.44  &7.88 &7.01 \\
\multicolumn{1}{l}{Mention replacement} & 7.07  &7.42 &6.54 \\
\multicolumn{1}{l}{Shuffle within segments} & 10.24  &12.32 &9.65 \\
\multicolumn{1}{l}{DAGA} &5.46 &6.23 &5.07 \\
\multicolumn{1}{l}{MELM} &5.27 &6.29 &4.82 \\
\multicolumn{1}{l}{\textbf{$\modelname$ (All)}}& \textbf{4.74} &\textbf{5.19} &\textbf{4.28} \\

\thickhline
\end{tabular}}
\caption{Perplexity of the augmented data with various augmentation approaches. Lower perplexity is better.}
\label{tab:perplexity}
\vspace{-1mm}
\end{table}

\noindent\textbf{Various Augmentation Multiples Performance}

We further vary the multiples of augmented data from 2x to 10x the training set to study the influence of data augmentation approaches for the NER backbone models under low resource scenarios. 
We choose different low resource datasets and three representative augmentation approaches (Mention replacement, MELM, and {\modelname} (All)), then represent the results in Figure \ref{fig:real}.

We could observe that the unified Seq2Seq framework has more performance gains with ever-increasing augmented data. {\modelname} (All) consistently achieves better F1 performance, with a clear margin, compared to baseline augmentation approaches under various augmentation multiples. Especially for Music, {\modelname} (All) brings an incredible 4.01\% improvement in F1 performance with only 300 augmented data.

\noindent\textbf{Semantic Coherence Analysis}

Compared with baseline augmentation approaches, {\modelname} conditionally generates texts with the diversity beam search decoding, which provides more coherent texts. We analyze the coherence through perplexity based on a large Transformer language model: GPT-2 \cite{radford2019language}. From Table \ref{tab:perplexity}, {\modelname} obtains the lowest perplexity. Although DAGA and MELM are also based on generative models, the texts are not natural enough since only partial text is replaced.
\begin{table}[t]
\centering
\scalebox{0.62}{
\begin{tabular}{lcccccc}
\thickhline
\multicolumn{1}{c}{\multirow{2}{*}{Methods / Datasets}} & \multicolumn{2}{c}{CoNLL2003} & \multicolumn{2}{c}{CADEC} &  \multicolumn{2}{c}{AI} 
\\ \cmidrule(lr){2-3} \cmidrule(lr){4-5} \cmidrule(lr){6-7}
&TTR & Diver.  &TTR  & Diver. &TTR & Diver.  \\
\midrule 
\multicolumn{1}{l}{Label-wise token rep.} & 81.2 & 3.1 & 80.5  &  3.4 & 81.9 & 3.3  \\
\multicolumn{1}{l}{Synonym replacement} & 81.9 & 3.3  & 80.1 & 3.5 & 82.6  & 3.4 \\
\multicolumn{1}{l}{Mention replacement} & \underline{83.8} & \underline{3.9} & \underline{82.9} & \underline{3.6} & \textbf{84.2} & \underline{3.8}\\
\multicolumn{1}{l}{Shuffle within segments} & 72.9 & 2.4 &71.6 &  2.0 & 73.7 & 2.1\\
\multicolumn{1}{l}{DAGA} & 73.8 & 2.8 & 74.1 & 2.6 & 74.3 & 3.1\\
\multicolumn{1}{l}{MELM} & 77.2 & 3.2 & 78.1 & 2.9 &76.6 & 3.0 \\
\multicolumn{1}{l}{\textbf{$\modelname$ (All)}} & \textbf{86.4} & \textbf{4.3} & \textbf{85.1} & \textbf{4.5} & \underline{83.7} & \textbf{4.4}\\

\thickhline
\end{tabular}}
\caption{Diversity Evaluation on three datasets.}
\label{tab:human}
\vspace{-1mm}
\end{table}

\noindent\textbf{Diversity Evaluation}

We measure the diversity of augmented sentences through automatic and manual metrics. For automatic metric, we introduce the Type-Token Ratio (TTR) \cite{tweedie1998variable} to evaluate the ratio of the number of different words to the total number for each original text. Higher TTR (\%) indicates more diversity in sentences. Besides that, we ask 5 annotators to give a score for the degree of diversity of the 200 generated texts, with score range of 1\textasciitilde5. According to the annotation guideline in Appendix \ref{guideline}, a higher score indicates the method can generate more diverse texts. 

We present the average scores on the datasets in Table \ref{tab:human}. {\modelname} could obtain 7.8\% TTR and 1.4 diversity performance boost in average compared to MELM.
\begin{table}[t!]
\centering
\resizebox{1.00\linewidth}{!}{
\begin{tabular}{l}
\toprule
\begin{tabular}[c]{@{}l@{}}Approach: {\color{blue}Original Sentence}
\\ 
Entity: \textbf{\texttt{unsupervised learning, principal component analysis}}\\
${\quad\quad}$\textbf{\texttt{, cluster analysis}}\\
Entity Type: \textbf{\texttt{field, algorithm, algorithm}}\\
Text: The main methods used in {\color{red}\textit{unsupervised learning}} are\\
${\quad\quad}$ {\color{red}\textit{principal component analysis}} and {\color{red}\textit{cluster analysis}}.
\end{tabular}     
\\\hline
\begin{tabular}[c]{@{}l@{}}Approach: {\color{blue}{\modelname} (Add)}
\\ 
Entity: \textbf{\texttt{unsupervised learning, principal component analysis}}\\
${\quad\quad}$\textbf{\texttt{, cluster analysis, dimension reduction}}\\
Entity Type: \textbf{\texttt{field, algorithm, algorithm, algorithm}}\\
Text: In {\color{red}\textit{unsupervised learning}}, {\color{red}\textit{principal component analysis}}, {\color{red}\textit{cluster analysis}} and\\
${\quad\quad}$ {\color{red}\textit{dimension reduction}} are used to reduce the number of variables in a task.
\end{tabular}     
\\
\hline
\begin{tabular}[c]{@{}l@{}}Approach: {\color{blue}{\modelname} (Delete)}
\\ 
Entity: \textbf{\texttt{unsupervised learning, principal component analysis}}\\
Entity Type: \textbf{\texttt{field, algorithm}}\\
Text: In the field of {\color{red}\textit{unsupervised learning}}, {\color{red}\textit{principal component analysis}}\\
${\quad\quad}$ is used to model the learning process.
\end{tabular}
\\
\hline
\begin{tabular}[c]{@{}l@{}}Operation: {\color{blue}{\modelname} (Replace)}
\\ 
Entity: \textbf{\texttt{unsupervised learning, principal component analysis}}\\
${\quad\quad}$ \textbf{\texttt{, dimension reduction}}\\
Entity Type: \textbf{\texttt{field, algorithm, algorithm}}\\
Text: In the field of {\color{red}\textit{unsupervised learning}}, {\color{red}\textit{principal component analysis}} and \\
${\quad\quad}${\color{red}\textit{dimension reduction}} are used to reduce the size of the data.
\end{tabular} \\
\hline
\begin{tabular}[c]{@{}l@{}}Operation: {\color{blue}{\modelname} (Swap)}
\\ 
Entity: \textbf{\texttt{unsupervised learning, cluster analysis}}\\
${\quad\quad}$\textbf{\texttt{, principal component analysis}}\\
Entity Type: \textbf{\texttt{field, algorithm, algorithm}}\\
Text: {\color{red}\textit{Unsupervised learning}} uses {\color{red}\textit{cluster analysis}} and {\color{red}\textit{principal component analysis}}\\
${\quad\quad}$ to learn a task.
\end{tabular} \\
\hline
\begin{tabular}[c]{@{}l@{}}Operation: {\color{blue}{\modelname} (All)}
\\ 
Entity: \textbf{\texttt{unsupervised learning, dimension reduction}}\\
${\quad\quad}$\textbf{\texttt{, principal component analysis}}\\
Entity Type: \textbf{\texttt{field, algorithm, algorithm}}\\
Text: {\color{red}\textit{Unsupervised learning}} uses {\color{red}\textit{cluster analysis}} to achieve the purpose of \\
${\quad\quad}${\color{red}\textit{dimension reduction}} for better learning a task.
\end{tabular} \\
\hline
\begin{tabular}[c]{@{}l@{}}Approach: {\color{blue}Mention Replacement}
\\ 
Entity: \textbf{\texttt{heterodyning, principal component analysis}}\\
${\quad\quad}$\textbf{\texttt{, cluster analysis}}\\
Entity Type: \textbf{\texttt{field, algorithm, algorithm}}\\
Text: The main methods used in {\color{red}\textit{heterodyning}} are {\color{red}\textit{principal component analysis}}\\
${\quad\quad}$ and {\color{red}\textit{cluster analysis}}.
\end{tabular}   \\
\hline
\begin{tabular}[c]{@{}l@{}}Operation: {\color{blue}{DAGA}}
\\ 
Text: {\color{red}\textit{Unsupervised learning}} uses {\color{red}\textit{principal component analysis}} and {\color{red}\textit{cluster analysis}}.\\
Entity (Unchanged): \textbf{\texttt{unsupervised learning, principal component}}\\
${\quad\quad}$\textbf{\texttt{analysis, cluster analysis}}\\
Entity Type (Unchanged): \textbf{\texttt{field, algorithm, algorithm}}\\
\end{tabular} \\
\bottomrule
\end{tabular}
}
\caption{The augmented texts for AI domain. We show six {\color{blue}approaches} to generate texts marked with the corresponding {\color{red}\textit{entity}} list.
}\label{tab:example_ai}
\vspace{-0.1in}
\end{table}

\section{Case Study}\label{case}

We show eight approaches to obtain augmented data for the AI domain in Table \ref{tab:example_ai}. Compared with baseline augmentation methods, {\modelname} introduces a knowledge expansion and conditionally generates texts based on the diversity beam search, which provides more coherent and diverse texts. For example, The Mention Replacement approach replaces the entities \texttt{unsupervised learning} with \texttt{heterodyning}, which ignores the semantics of the context and makes an ungrammatical replacement, resulting in incoherent and unreasonable texts. For the DAGA approach, it simply stacks three entities: \texttt{unsupervised learning, principal component analysis, cluster analysis} in the text, which could not provide knowledge expansions to the NER models.

\section{Conclusions and Future Work}
\label{sec:conclusion}
In this paper, we propose an Entity-to-Text based data augmentation approach {\modelname} for NER tasks. 
Compared with traditional rule-based augmentation methods that break semantic coherence, or use Text-to-Text based augmentation methods that cannot be used on nested and discontinuous NER tasks, our method can generate semantically coherent texts for all NER tasks, and use the diversity beam search to improve the diversity of augmented texts. Experiments on thirteen public real-world datasets, and coherence and diversity analysis show the effectiveness of {\modelname}. Moreover, we can also apply the method of data augmentation to low-resource relation extraction \cite{hu2020selfore,hu2021gradient,hu2021semi,liu2022hierarchical,hu2023think}, natural language inference \cite{li2023multi,li2022pair}, semantic parsing \cite{liu2022semantic,liu2023comprehensive}, and other NLP application tasks, thus realizing knowledge enhancement based on data augmentation approach.

\section{Limitations}
We discuss the limitations of our method from three perspectives. 

First, our method is based on pre-trained language models, so compared to rule-based data augmentation methods (synonym replacement, shuffle within segments, etc.), our method requires higher time complexity. 

Second, the entity matching process (Section \ref{exploitation}) will discard sentences which cannot match entities in the entity list, which will affect the utilization of data. 

Third, our data augmentation method based on the pre-trained language models, whose generalization ability is limited since the augmented knowledge comes from the pre-trained language models. However, the knowledge in pre-trained language models is limited and not domain-specific. How to improve the generalization ability of the data augmentation methods is a future research work.

\section{Acknowledgement}
We thank the reviewers for their valuable comments. Yong Jiang and Lijie Wen are the corresponding authors. Xuming Hu, Aiwei Liu and Lijie Wen were partially supported by the National Key Research and Development Program of China (No. 2019YFB1704003), the National Nature Science Foundation of China (No. 62021002), Tsinghua BNRist and Beijing Key Laboratory of Industrial Bigdata System and Application. Philip S. Yu was partially supported by the NSF under grants III-1763325, III-1909323, III-2106758, SaTC-1930941.

\bibliography{custom}
\bibliographystyle{acl_natbib}

\clearpage
\newpage
\appendix
\begin{table*}[t!]
\centering
\scalebox{0.75}{
\begin{tabular}{ccccccccccc}
\toprule
\multicolumn{2}{c}{\multirow{2}{*}{}} & \multicolumn{5}{c}{Sentence} & \multicolumn{4}{c}{Entity} 
\\ \cmidrule(lr){3-7} \cmidrule(lr){8-11}
& & \#All & \#Train & \#Dev & \#Test & \#Avg.Len & \#All & \#Nes. & \#Dis. & \#Avg.Len \\
\midrule 
\multicolumn{1}{c}{\multirow{7}{*}{Flat NER}} &CoNLL2003 &20,744 &17,291 &-- &3,453 &14.38 &35,089 &-- &-- &1.45 \\
&OntoNotes &76,714 &59,924 &8,528 &8,262 &18.11 &104,151 &-- &-- &1.83 \\\cmidrule(lr){2-11}
&Politics &1,392 &200 &541 &651 &50.15 &22,854 &-- &-- &1.35 \\
&Nature Science &1,193 &200 &450 &543 &46.50 &14,671 &-- &-- &1.72 \\
&Music &936 &100 &380 &456 &48.40 &15,441 &-- &-- &1.37 \\
&Literature &916 &100 &400 &416 &45.86 &11,391 &-- &-- &1.47 \\
&AI &881 &100 &350 &431 &39.57 &8,260 &-- &-- &1.55 \\
\midrule 
\multicolumn{1}{c}{\multirow{3}{*}{Nested NER}} &ACE2004 &8,512 &6,802 &813 &897 &20.12 &27,604 &12,626 &-- &2.50 \\
&ACE2005 &9,697 &7,606 &1,002 &1,89 &17.77 &30,711 &12,404 &-- &2.28 \\
&Genia &18,546 &15,023 &1,669 &1,854 &25.41 &56,015 &10,263 &-- &1.97 \\
\midrule 
\multicolumn{1}{c}{\multirow{3}{*}{Discontinuous NER}} &CADEC &7,597 &5,340 &1,097 &1,160 &16.18 &6,316 &920 &670 &2.72 \\
&ShARe13 &18,767 &8,508 &1,250 &9,009 &14.86 &11,148 &663 &1,088 &1.82 \\
&ShARe14 &34,614 &17,404 &1,360 &15,850 &15.06 &19,070 &1,058 &1,656 &1.74 \\

\bottomrule
\end{tabular}}
\caption{Dataset statistics. ``\#'' denotes the amount. ``Nes.'' and ``Dis.'' denote nested and discontinuous entities respectively.}
\label{tab:dataset_sta}
\end{table*}
\section{Dataset Statistics}\label{Dataset}
we show the detailed statistics of the datasets in Table \ref{tab:dataset_sta}. We further give details on entity types for thirteen datasets in Table \ref{tab:real low dataset}.

\begin{table}[h]
\centering
\caption{Detailed statistics on entity types for thirteen NER datasets.}
\resizebox{0.99\linewidth}{!}{
\begin{tabular}{cc}
\toprule
Datasets                                                                   & Entity Types                                                     \\ \hline
\multirow{1}{*}{CoNLL2003}                                                  & location, organization, person, miscellaneous,                          \\ \hline
\multirow{3}{*}{OntoNotes}                                                  & person, norp, facility, organization, gpe,        \\
                                                                           & location, product, event, work of art, law, language  \\
                                                                           & date, time, percent, money, quantity, ordinal, cardinal
                                                                           \\ \hline
\multirow{2}{*}{ACE2004}                                                  & gpe, organization, person, facility,        \\
                                                                           & vehicle, location, wea                        \\ \hline
\multirow{2}{*}{ACE2005}                                                  & gpe, organization, person, facility,       \\
                                                                           & vehicle, location, wea                        \\ \hline
\multirow{1}{*}{Genia}                                                  & protein, cell\_type, cell\_line, RNA,  DNA,                        \\ \hline
\multirow{1}{*}{CADEC}                                                  &  ade                       \\ \hline
\multirow{1}{*}{ShARe13}                                                  &   disorder                 \\ \hline                                       
\multirow{1}{*}{ShARe14}                                                  &   disorder                  \\ \hline

\multirow{2}{*}{Politics}                                                  & politician, person, organization, political party, event,        \\
                                                                           & election, country, location, miscellaneous                       \\ \hline
\multirow{4}{*}{\begin{tabular}[c]{@{}c@{}}Natural\\ Science\end{tabular}} & scientist, person, university, organization, country,            \\
                                                                           & enzyme, protein, chemical compound, chemical element,            \\
                                                                           & event, astronomical object, academic journal, award,             \\
                                                                           & location, discipline, theory, miscellaneous                      \\ \hline
\multirow{3}{*}{Music}                                                     & music genre, song, band, album, musical artist,                  \\
                                                                           & musical instrument, award, event, country,                       \\
                                                                           & location, organization, person, miscellaneous                    \\ \hline
\multirow{2}{*}{Literature}                                                & writer, award, poem, event, magazine, person, location,          \\
                                                                           & book, organization, country, miscellaneous                       \\ \hline
\multirow{2}{*}{AI}                                                        & field, task, product, algorithm, researcher, metrics, university \\
                                                                           & country, person, organization, location, miscellaneous           \\ \bottomrule
\end{tabular}}
\label{tab:real low dataset}
\end{table}

\section{Entity Addition and Replacement Strategy}

{\modelname} add and replace the entity in the training set that has the same entity type. This strategy can provide the knowledge expansion during the generation, which is an appealing property when the hand-craft knowledge base is difficult to construct for augmentation approaches. 

If we directly replace the entities in the text with other entities of the same type, this is equivalent to the baseline: Mention Replacement. From Table \ref{tab:verification}, \ref{tab:low-resource} and \ref{tab:real}, we could observe that compared to {\modelname} (Replace), the improvement of F1 performance is greatly reduced. The main reason is that context-free entities are replaced, resulting in obscure and unreasonable texts. For example, ``\textit{EU’s \underline{German} wing says it has received a warning}.'' may be changed to ``\textit{EU’s \underline{World War Two} wing says it has received a warning}.'' since the two entities share the same type: MISC.

\section{Hyperparameter Analysis}\label{hyperparmeter}

We study the hyperparameter $\gamma$ in the diversity beam search, which represents the degree of probability penalty in the decoding process and determines the diversity of sentences. Modifying $\gamma$ allows us to control the diversity of the texts. We vary the $\gamma$ from 1 to 100 and represent the F1 results using the unified Seq2Seq framework and {\modelname} (All) in Table \ref{tab:Hyperparameters}. With no more than 1\% F1 fluctuating results among three datasets, {\modelname} appears robust to the choice of $\gamma$.

\begin{table}[t!]
\centering
\resizebox{0.97\linewidth}{!}{
\begin{tabular}{lcccccc}
\toprule
Datasets / ${\gamma}$ & 1 & 5  & 10  & 25 & 50 & 100 \\
\midrule 
\multicolumn{1}{l}{CoNLL2003} &93.01  &93.26 &\textbf{93.51} &93.44 &93.28 &93.16 \\
\multicolumn{1}{l}{ACE2005} &85.46  &\textbf{86.41} &86.39 &86.30 &86.06 &85.77\\
\multicolumn{1}{l}{CADEC} &70.88  &71.34 &\textbf{71.70} &71.64 &71.42 &70.99\\
\bottomrule
\end{tabular}}
\caption{F1 results under different ${\gamma}$ using the unified Seq2Seq framework and {\modelname} (All).}
\label{tab:Hyperparameters}
\vspace{-0.1in}
\end{table}

\section{Annotation Guideline}\label{guideline}
Each annotator needs to carefully read each augmented text, compare it with the original text, and give a score according to the following criteria. Note that all augmented texts for a dataset are given an average score.
\begin{itemize}
    \item Score:1. The augmented texts under the same original text are almost the same.
    \item Score:2. The augmented texts under the same original text are slightly different, with serious grammatical errors.
    \item Score:3. The augmented texts under the same original text are slightly different, and there are almost no grammatical errors.
    \item Score:4. The augmented texts under the same original text are diverse, with serious grammatical errors.
    \item Score:5. The augmented texts under the same original text are diverse, and there are almost no grammatical errors.
\end{itemize}

\end{document}